\documentclass[conference]{IEEEtran}
\IEEEoverridecommandlockouts
\usepackage{cite}
\usepackage{amsmath,amssymb,amsfonts}
\usepackage{algorithmic}
\usepackage{graphicx}
\usepackage{textcomp}
\usepackage{xcolor}
\usepackage{booktabs}
\usepackage{multirow}
\usepackage{hyperref}
\usepackage{float}
\restylefloat{table}
\usepackage{graphbox}
\usepackage{multirow}% it loads the graphicx package

\usepackage{subcaption}
\def\BibTeX{{\rm B\kern-.05em{\sc i\kern-.025em b}\kern-.08em
    T\kern-.1667em\lower.7ex\hbox{E}\kern-.125emX}}
\begin{document}

\title{Domain Adaptable Self-supervised Representation Learning on Remote Sensing Satellite Imagery \\
}

\author{
    \IEEEauthorblockN{
        Muskaan Chopra\textsuperscript{$\gamma$, *},
        Prakash Chandra Chhipa\textsuperscript{$\delta$, *},
        Gopal Mengi\textsuperscript{$\gamma$, *},
        Varun Gupta\textsuperscript{$\gamma$} and
        Marcus Liwicki\textsuperscript{$\delta$}\\
        \textit{\textsuperscript{$\delta$} Machine Learning Group, EISLAB, Lule\aa~Tekniska Universitet, Lule\r{a}, Sweden}\\
        \textit{\{prakash.chandra.chhipa, marcus.liwicki\}@ltu.se}\\
        \textit{\textsuperscript{$\gamma$} Chandigarh College of Engineering and Technology, Punjab University, Chandigarh, India}\\
        \textit{chopramuskaan47@gmail.com, gopalmengi@gmail.com, varungupta@ccet.ac.in}\\
        \textit{\textsuperscript{*}  co-first authors with equal contribution}
     }
    }

% \and
% \and
% \and
% \and
% \and

\maketitle

\begin{abstract}
This work presents a novel domain adaption paradigm for studying contrastive self-supervised representation learning and knowledge transfer using remote sensing satellite data. Major state-of-the-art remote sensing visual domain efforts primarily focus on fully supervised learning approaches that rely entirely on human annotations. On the other hand, human annotations in remote sensing satellite imagery are always subject to limited quantity due to high costs and domain expertise, making transfer learning a viable alternative. The proposed approach investigates the knowledge transfer of self-supervised representations across the distinct source and target data distributions in depth in the remote sensing data domain. In this arrangement, self-supervised contrastive learning- based pretraining is performed on the source dataset, and downstream tasks are performed on the target datasets in a round-robin fashion. Experiments are conducted on three publicly available datasets, UC Merced Landuse (UCMD), SIRI-WHU, and MLRSNet, for different downstream classification tasks versus label efficiency. In self-supervised knowledge transfer, the proposed approach achieves state-of-the-art performance with label efficiency labels and outperforms a fully supervised setting. A more in-depth qualitative examination reveals consistent evidence for explainable representation learning. The source code and trained models are published on GitHub\footnote{\href{https://github.com/muskaan712/Domain-Adaptable-Self-Supervised-Representation-Learning-on-Remote-Sensing-Satellite-Imagery} {https://github.com/muskaan712/Domain-Adaptable-Self-Supervised-Representation-Learning-on-Remote-Sensing-Satellite-Imagery}}.
\end{abstract}


\section{Introduction}
\begin{figure}[htbp]
\centerline{
\includegraphics[width=9cm, height=4cm]{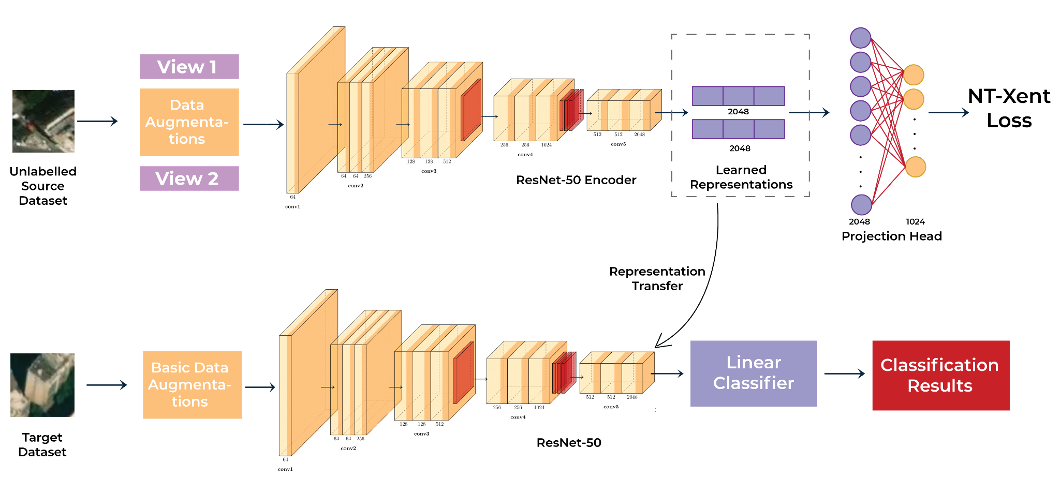} 
}
\caption{Demonstrate the instance of proposed domain adaptation framework where self-supervised representation learning is performed at source dataset using contrastive learning method and downstream task performed on target datasets.}
\label{fig2}
\end{figure}
To formulate the policies and schemes, the region's geographical and demographic information and its efficient representation are essential \cite{b1}\cite{b2}. Visual interpretation of aerial and space images is the most common method of producing topographic and thematic maps. Satellite images are also used to classify different types of crops using deep learning techniques \cite{b3}\cite{b4}. Today many high-resolution satellites can be relied upon to develop cartographic projects \cite{b5}. But it's not always the case when you get a high-resolution image which makes a major concern. Satellite images are not always provided in abundance, and there may be fewer image samples, which further poses a challenge in classification and segmentation. The applications of satellite imagery classification include disaster prediction using remote sensing images, and these early predictions are used to take necessary precautions.\cite{b31} Satellite images can also be classified and segmented into areas with more wind and solar power so that adequate coverage of windmills and solar panels can be achieved to harness the power efficiently.\cite{b32}\\
Substantial human-labeled data is necessary to train a deep neural network successfully. Unfortunately, data collection and labeling are time-consuming and challenging in many fields. However, acquiring sufficient annotated data can be quite expensive and time-intensive. The process of cleaning, screening, labeling, evaluating, and reorganizing data by a training framework can be exceedingly time-consuming and complicated \cite{b11}. The lack of data has spawned a variety of solutions, the most prevalent of which is transfer learning. This work presents an approach that minimizes the training samples and puts less stress on the data labeling compared to the architecture modeling \cite{b6}. For most supervised learning approaches, annotated data is necessary to train a machine. \\
This work employs self-supervised learning-based satellite image classification to deal with scarce labeled data in satellite imagery. When representations are learned from a pretext task using unlabeled input images and then used for a downstream task of interest, self-supervised learning is akin to transfer learning \cite{b12}. This work has used Domain Adaptation (DA) to prove the robustness of the model toward performance generalization on unseen data distribution.
Domain Adaptation follows the concept that the model gets trained on one source dataset and evaluated on the other target dataset, increasing the model's reliability, re-usability, and results. 
%Reliability is a crucial factor as the performance of applications of such methods is required for government policies, and any mismatched information may compromise the decision-making process\cite{b10}. The %generalizability and accuracy of developed solutions are hampered because most research institutions and enterprises only have limited access to annotated satellite images. 
%To the best of our knowledge, this paper proposes domain adaptable self-supervised learning-based satellite image classification for the first time in literature. 
The proposed work's main contributions are outlined as-
\begin{itemize}
\item Establishing the adaptation of the self-supervised representation learning on remote-sensing satellite imagery by proposing a domain adaptation framework for rigorous evaluation.
\item Achieving label-efficient representational knowledge transfer across multiple public datasets by obtaining state-of-the-art performance with limited labels and outperforming in fully supervised settings.
\item Explaining improvement in quantitative results by qualitative analysis with significant and consistent evidence.
%\iten We propose self-supervised representation learning-based classification of satellite imagery which requires less labeled data.
%\item The proposed method of satellite image classification is domain adaptable. The model trained on one dataset of satellite images to learn image representations can classify satellite images  from different datasets with different classes.  
%\item The proposed method outperforms the existing ones and obtains state-of-the-art results for satellite image classification tasks. 
\end{itemize}
The rest of the article is organized as follows. Section 2 discusses the datasets. Section 3 presents a domain adaptation framework for self-supervised contrastive learning. Section 4 presents the experiments and results. Section 5 discusses the experiments and results obtained from the proposed work, Section 6 presents the related work, and Section 7 concludes the proposed work and provides the future scope of the work. 
%Source code and trained models are available in GitHub repository\footnote{https://github.com/muskaan712/Domain-Adaptable-Self-Supervised-Representation-Learning-on-Remote-Sensing-Satellite-Imagery}.
\section{Satellite Imagery Dataset Description}
There are many applications for satellite images in meteorology, oceanography, fisheries, agriculture, biodiversity, geology, cartography, and land use planning. Instead of only having an image of a place, satellite image classification aims to transform satellite imagery into valuable information. Satellite Imagery of residential and non-residential buildups varies with objects and natural scenes captured in the image. A dataset with images labeled as a whole is required for categorizing satellite images. Three public satellite imagery datasets are used in this work, SIRI-WHU and UC Merced have equally distributed images among their classes and MLRSNet has non-uniform distribution which is demostrated in the graph below, other details about the datasets are discussed below and summarized in Table \ref{tab1}.
\begin{table}[htbp]
\caption{Dataset description}
\begin{center}
\begin{tabular}{|p{2.5cm}|c|c|}
\hline
\textbf{\textit{Dataset}}& \textbf{\textit{Total Images}}& \textbf{\textit{No. of Classes}} \\
\hline
\textit{SIRI-WHU}& \textit{2400}& \textit{12} \\
\hline
\textit{UC Merced Land Use Dataset}& \textit{ 2100}& \textit{21} \\
\hline
\textit{MLRSNet}& \textit{109,161}& \textit{46} \\
\hline
\end{tabular}
\label{tab1}
\end{center}
\end{table}
\begin{figure}[htbp]
\centerline{
\includegraphics[width=9cm, height=6cm]{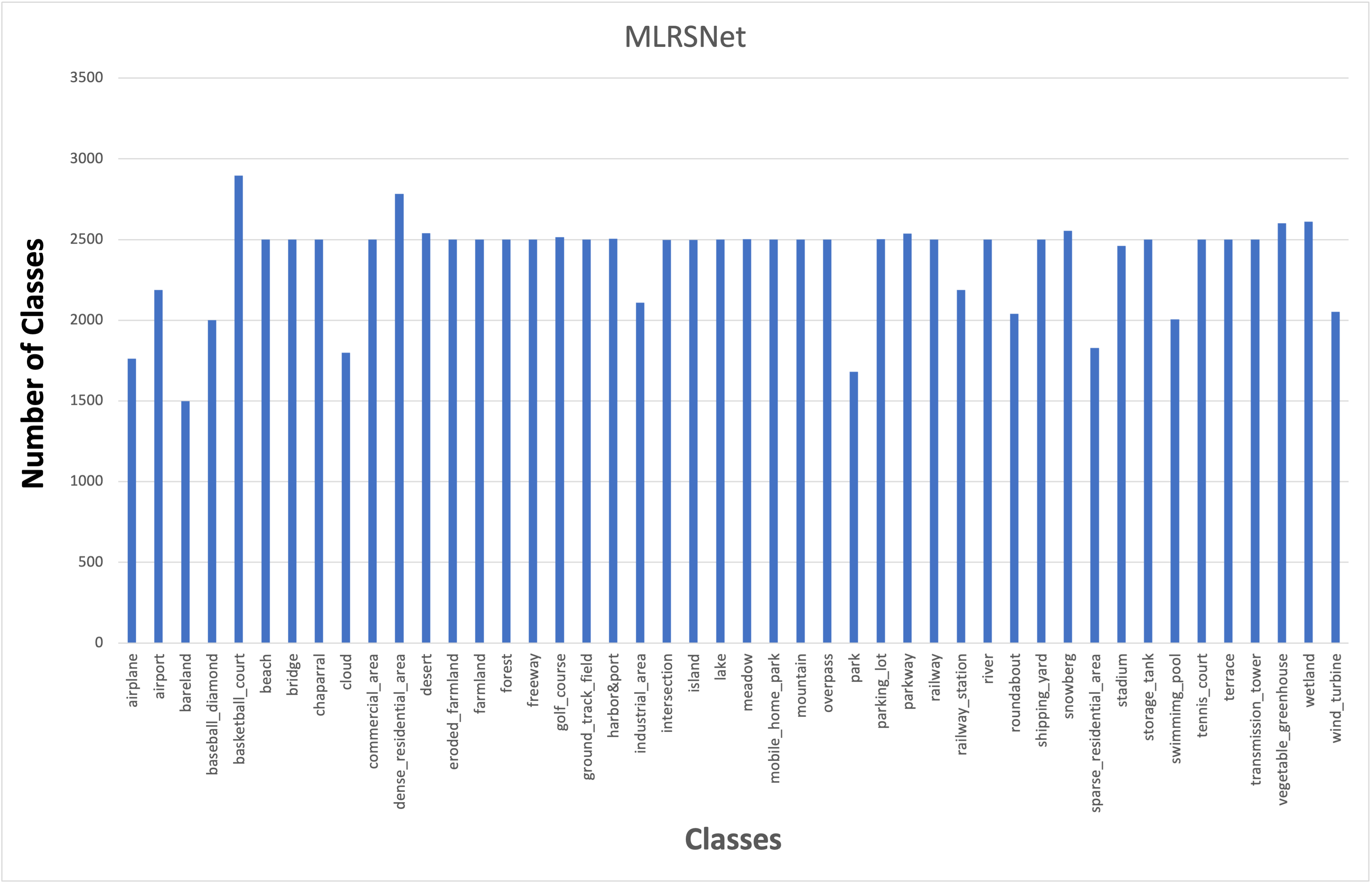} 
}
\caption{Class-wise distribution of the MLRSNet dataset.}
\label{fig1}
\end{figure}
\subsection{SIRI-WHU Dataset}
The SIRI-WHU dataset\footnote{\href{http://www.lmars.whu.edu.cn/prof_web/zhongyanfei/e-code.html}{http://www.lmars.whu.edu.cn/prof\_web/zhongyanfei/e-code.html}} for classification has 2400 photos sorted into 12 classifications. This dataset was obtained from Google Earth and mainly included metropolitan regions in China, with the image collection developed by Wuhan University's RS IDEA Group (SIRI-WHU). It consists of 12 classes: Agriculture, Commercial, Harbor, Idle land, Industrial, Meadow, Overpass, Park, Pond, Residential, River, and Water. Each class comprises 200 pictures that are 200 x 200 pixels in size. 
\subsection{UC Merced Dataset}
The image data in the UC Merced dataset\footnote{\href{http://weegee.vision.ucmerced.edu/datasets/landuse.html}{http://weegee.vision.ucmerced.edu/datasets/landuse.html}} were manually extracted from large-sized images in the United States Geological Survey (USGS) National Map Urban Area Imagery collection for numerous cities across the country (United States). This big ground truth picture collection consists of 21 land-use types, each with 100 pictures. The 21 classes were namely agricultural, airplane, baseball diamond, beach, buildings, chaparral, dense residential, forest, freeway, golf course, harbor, intersection, medium residential, mobile home park, overpass, parking lot, river, runway, sparse residential, storage tanks, and tennis court. This public domain imagery has a pixel resolution of 1 foot, with each image being 256x256 pixels. 
\subsection{MLRSNet}
MLRSNet\footnote{\href{https://data.mendeley.com/datasets/7j9bv9vwsx/2}{https://data.mendeley.com/datasets/7j9bv9vwsx/2}} offers several satellite-based perspectives of the world. It comprises optical satellite images with great spatial resolution—between 1,500 and 3,000 example photos in every 46 categories in the 109,161 remote sensing photographs makeup MLRSNet. The photos are 256256 pixels and have different pixel sizes (10m to 0.1m). The dataset can be used for picture segmentation, image retrieval, and classification based on multiple labels.
% \begin{figure}[htbp]
% \centerline{
% \begin{subfigure}{0.075\textwidth}
% \includegraphics[width=1.35cm, height=1.35cm]{figure1a.jpg} 
% \caption{}
% \label{fig:subim1}
% \end{subfigure}
% \begin{subfigure}{0.08\textwidth}
% \includegraphics[width=1.35cm, height=1.35cm]{figure1b.jpg}
% \caption{}
% \label{fig:subim2}
% \end{subfigure}
% \begin{subfigure}{0.075\textwidth}
% \includegraphics[width=1.35cm, height=1.35cm]{figure1c.jpg}
% \caption{}
% \label{fig:subim2}
% \end{subfigure}
% \begin{subfigure}{0.08\textwidth}
% \includegraphics[width=1.35cm, height=1.35cm]{figure1d.jpg}
% \caption{}
% \label{fig:subim2}
% \end{subfigure}
% \begin{subfigure}{0.075\textwidth}
% \includegraphics[width=1.35cm, height=1.35cm]{figure1e.png}
% \caption{}
% \label{fig:subim2}
% \end{subfigure}
% \begin{subfigure}{0.08\textwidth}
% \includegraphics[width=1.35cm, height=1.35cm]{figure1f.png}
% \caption{}
% \label{fig}
% \end{subfigure}
% }
% \caption{Sample images- (a)(b) subset of EyePACS Dataset, (c)(d) Messidor-I dataset,
% (e)(f) APTOS 2019 dataset.}
% \label{fig}
% \end{figure}
\section{Domain Adaptation framework for Self-supervised contrastive learning}
The proposed framework consists of two main tasks: (i) pretext task, in which learning representations following contrastive self-supervised learning on satellite imagery datasets within the source domain is performed (ii) downstream task, in which satellite images are classified based on the representations learned in pretext task. Figure \ref{fig1} depicts a schematic diagram of the proposed approach where knowledge transfer on self-supervised learnt representation is comprehensively validated\\
\begin{figure}[htbp]
\centerline{
\includegraphics[width=8cm, height=9cm]{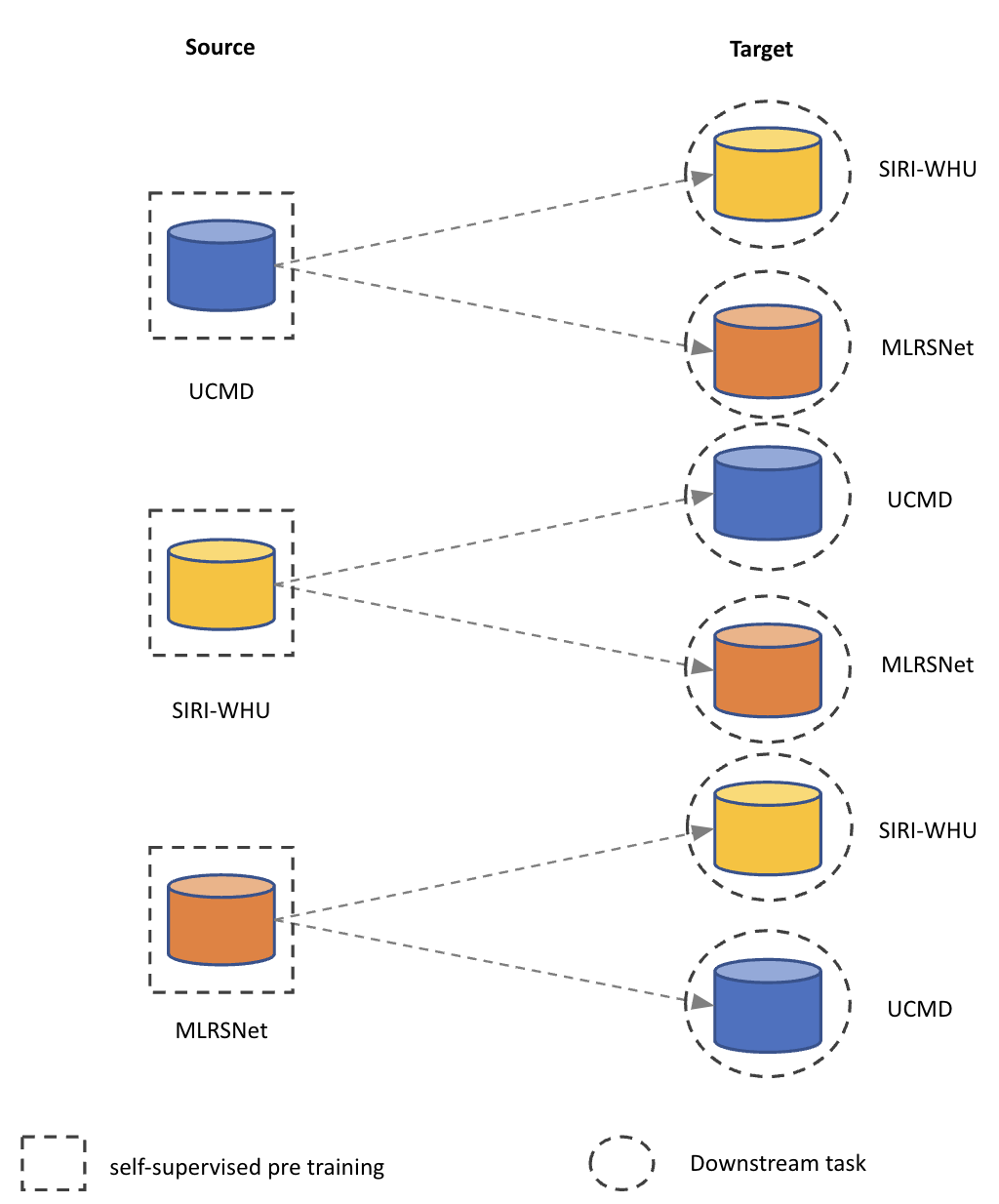} 
}
\caption{Proposed approach for domain adaptation. It ensures to investigate of each dataset for self-supervised representation learning and for downstream tasks under all possible domain adaptation scenarios}
\label{fig1}
\end{figure}
In the pretext task, various augmentations are applied to the images, such as flipping, affine transformations, jitter, grayscale, etc., to create different views of the images.
A positive pair is a pair of views created from the same image, whereas a negative pair is a pair of views created from different images. Then, positive and negative pairs of images are contrastively learned to form image representations. Labeled images are not required for representation learning in this task.\\
Figure \ref{fig2} depicts the contrastive learning architecture. 
SimCLR (simple framework for contrastive learning)\cite{b21} has been used to learn the representations. Positive and negative pairs of satellite images are created from unlabelled satellite images with augmentations such as Gaussian blur, flipping, translation, rotation, etc. These pairs of positive and negative views are fed to the encoder network. ResNet-50 encoder is a backbone for the pretext task network, followed by two fully connected layers containing 2048 and 1024 neurons each. The encoder part helps in extracting image representations for positive and negative pairs.
Normalized Temperature-scaled Cross-Entropy loss (NT-Xent) is used to pull close representations and push away different representations. This loss function for a positive pair is defined below.
$$
\ell\left(\mathbf{z}_{\mathrm{i}}, \mathbf{z}_{\mathrm{j}}\right)=-\log \frac{\exp \left(\operatorname{sim}\left(z_i z_j\right) / \tau\right)}{\sum_{k=1}^{2 n} 1_{k \neq i} \exp \left(z_i z_k / \tau\right)}
$$
T is the temperature parameter, where $z_{i}$ and $z_{j}$ in the numerator represent positive pairs, where $z_{i}$ and $z_{k}$  in the denominator represent all possible pairs, including positive and negative.
In terms of a loss function, it comes down to the ratio of the sum of similarities between all positive pairs divided by the negative log-likelihood of these pairs being similar. A softmax function-based temperature parameter is used to normalize this loss function. It is designed to maximize an agreement between positive pairs in a mini-batch.
The loss function for all the positive pairs is given below.
$$
\mathcal{L}=-\frac{1}{N} \sum_{i, j \in \mathcal{M} \mathcal{B}} \log \frac{\exp \left(\operatorname{sim}\left(z_i, z_j\right) / \tau\right)}{\sum_{k=1}^{2 N} \mathbb{1}_{[k \neq i]} \exp \left(\operatorname{sim}\left(z_i, z_k\right) / \tau\right)}
$$
The downstream task uses the learned embeddings of images during the pretext task as input. Only a few basic augmentations like resizing and cropping are used during the downstream task. The downstream task involves binary and multi-class classification of satellite images. A binary and multi-class classification task is involved in the downstream task. For classification, fewer labeled images are now required for the downstream task.
In the downstream task, the input image from the target dataset is taken as input, and primary augmentations (resizing, cropping) are applied to this image. The augmented images are fed to the encoder, initialized from the pretext task-trained model. A linear classifier having layers 512 and a number of classes has been appended to the encoder part of the network to classify satellite images.

\section{Experiments and Results}

Extensive experimentation is designed and performed to investigate the domain adaptation in self-supervised learning based representational knowledge transfer on three datasets, UC Merced, SIRI-WHU, and MLRSNet, covering binary and multi-class classifications downstream tasks under varying label efficiency.   
%We have performed various experiments for performing binary and multi-classification of satellite images taken from a number of datasets such as UC Merced, SIRI-WHU, and MLRSNet.
Table \ref{tab2} shows the augmentations for the pretext task, and Table \ref{tab2_hp} shows the hyperparameters used for training the pretext task. Details of hyperparameters for fine-tuning and other configuration is available in open-source source code. The dataset follows a 70\%, 20\%, and 10\% split for training, testing, and validation. The next subsections discuss the binary classification results and the multi-class classification results. The performance metrics are defined below.
$$
\text { Precision }=\frac{\text { Total true positives }}{\text { Real actual positives }+\text { Total false positives }}
$$
$$
\text { Recall }=\frac{\text { Total true positives }}{\text { Total true positives }+\text { Total false-negatives }}
$$
$$
\text { Accuracy }=\frac{\text { true negatives }+\text { true positives }}{\text { total cases }}
$$
$$
\text { F1 Score }=2 * \frac{\text { Precision } * \text { Recall }}{\text { Precision }+\text { Recall }}
$$

\begin{table}[htbp]
\caption{Augmentations for pretext task}
\begin{center}
\begin{tabular}{|c|c|}
\hline
\textbf{\textit{Augmentations}}& \textbf{\textit{Value}} \\
\hline
\textit{Resize}& \textit{224 $\times$ 224} \\
\hline
\textit{Horizontal Flip}& \textit{P= 0.5} \\
\hline
\textit{Vertical Flip}& \textit{P - 0.5} \\
\hline
\textit{Rotation}& \textit{(-90, 90)} \\
\hline
\textit{Grayscale}& \textit{P - 0.2} \\
\hline
\textit{Gaussian Blur}& \textit{P - 0.51, Kernel size - [21, 21]} \\
\hline
\end{tabular}
\label{tab2}
\end{center}
\end{table}

\begin{table}[htbp]
\caption{Hyperparameters for pretext task}
\begin{center}
\begin{tabular}{|c|c|}
\hline
\textbf{\textit{Hyperparameters}}& \textbf{\textit{Value}} \\
\hline
\textit{Batch size}& \textit{256} \\
\hline
\textit{Optimizer}& \textit{SGD} \\
\hline
\textit{Momentum}& \textit{0.9, nesterov=True} \\
\hline
\textit{Learning Rate}& \textit{0.0005} \\
\hline
\textit{Weight decay}& \textit{0.0005} \\
\hline
\end{tabular}
\label{tab2_hp}
\end{center}
\end{table}

\subsection{Domain Adaptation }
For domain adaptation, three different datasets have been used to evaluate the results and performance of the proposed methodology for satellite image classification. The three datasets are used without labels in the pretext task to generate representations and fine-tuning is performed on the other two remaining datasets to evaluate domain adaptation.  The results of experiments are shown in Table \ref{tab3}.
          
\begin{table}[htbp]
\caption{Results for Domain Adaptation on multiclassification}
\label{tab3}
\resizebox{\columnwidth}{!}{%
\begin{tabular}{|lll|c|c|c|c|}
\hline
\multicolumn{3}{|c|}{\textit{\textbf{Dataset used}}}                                                                                                                                                        & \multirow{2}{*}{\textit{\textbf{Accuracy}}} & \multirow{2}{*}{\textit{\textbf{Precision}}} & \multirow{2}{*}{\textit{\textbf{Recall}}} & \multirow{2}{*}{\textit{\textbf{F1-Score}}} \\ \cline{1-3}
\multicolumn{1}{|l|}{\textit{\textbf{Pretext}}}           & \multicolumn{1}{l|}{\textit{\textbf{Downstream}}}        & \textit{\textbf{\begin{tabular}[c]{@{}l@{}}\% of data in \\ downstream\end{tabular}}} &                                             &                                              &                                           &                                             \\ \hline
\multicolumn{1}{|l|}{\multirow{3}{*}{\textit{UC Merced}}} & \multicolumn{1}{l|}{\multirow{3}{*}{\textit{MLRSNet}}}   & \textit{100\%}                                                                        & \textit{96.34}                              & \textit{96.21}                               & \textit{96.56}                            & \textit{96.87}                              \\ \cline{3-7} 
\multicolumn{1}{|l|}{}                                    & \multicolumn{1}{l|}{}                                    & \textit{50\%}                                                                         & \textit{95.18}                              & \textit{94.83}                               & \textit{94.76}                            & \textit{94.34}                              \\ \cline{3-7} 
\multicolumn{1}{|l|}{}                                    & \multicolumn{1}{l|}{}                                    & \textit{10\%}                                                                         & \textit{92.23}                              & \textit{91.98}                               & \textit{91.73}                            & \textit{91.45}                              \\ \hline
\multicolumn{1}{|l|}{\multirow{3}{*}{\textit{UC Merced}}} & \multicolumn{1}{l|}{\multirow{3}{*}{\textit{SIRI-WHU}}}  & \textit{100\%}                                                                        & \textit{96.87}                              & \textit{96.32}                               & \textit{96.34}                            & \textit{96.87}                              \\ \cline{3-7} 
\multicolumn{1}{|l|}{}                                    & \multicolumn{1}{l|}{}                                    & \textit{50\%}                                                                         & \textit{94.99}                              & \textit{94.12}                               & \textit{94.76}                            & \textit{94.12}                              \\ \cline{3-7} 
\multicolumn{1}{|l|}{}                                    & \multicolumn{1}{l|}{}                                    & \textit{10\%}                                                                         & \textit{87.50}                              & \textit{87.43}                               & \textit{87.24}                            & \textit{87.97}                              \\ \hline
\multicolumn{1}{|l|}{\multirow{3}{*}{\textit{MLRSNet}}}   & \multicolumn{1}{l|}{\multirow{3}{*}{\textit{UC Merced}}} & \textit{100\%}                                                                        & \textit{98.50}                              & \textit{98.54}                               & \textit{98.21}                            & \textit{98.32}                              \\ \cline{3-7} 
\multicolumn{1}{|l|}{}                                    & \multicolumn{1}{l|}{}                                    & \textit{50\%}                                                                         & \textit{96.01}                              & \textit{96.80}                               & \textit{96.79}                            & \textit{96.85}                              \\ \cline{3-7} 
\multicolumn{1}{|l|}{}                                    & \multicolumn{1}{l|}{}                                    & \textit{10\%}                                                                         & \textit{92.32}                              & \textit{92.32}                               & \textit{92.56}                            & \textit{92.81}                              \\ \hline
\multicolumn{1}{|l|}{\multirow{3}{*}{\textit{MLRSNet}}}   & \multicolumn{1}{l|}{\multirow{3}{*}{\textit{SIRI-WHU}}}  & \textit{100\%}                                                                        & \textit{97.50}                              & \textit{97.11}                               & \textit{96.98}                            & \textit{96.43}                              \\ \cline{3-7} 
\multicolumn{1}{|l|}{}                                    & \multicolumn{1}{l|}{}                                    & \textit{50\%}                                                                         & \textit{96.24}                              & \textit{96.87}                               & \textit{96.32}                            & \textit{96.76}                              \\ \cline{3-7} 
\multicolumn{1}{|l|}{}                                    & \multicolumn{1}{l|}{}                                    & \textit{10\%}                                                                         & \textit{89.58}                              & \textit{89.90}                               & \textit{90.45}                            & \textit{89.34}                              \\ \hline
\multicolumn{1}{|l|}{\multirow{3}{*}{\textit{SIRI-WHU}}}  & \multicolumn{1}{l|}{\multirow{3}{*}{\textit{UC Merced}}} & \textit{100\%}                                                                        & \textit{98.75}                              & \textit{98.21}                               & \textit{97.93}                            & \textit{98.53}                              \\ \cline{3-7} 
\multicolumn{1}{|l|}{}                                    & \multicolumn{1}{l|}{}                                    & \textit{50\%}                                                                         & \textit{96.51}                              & \textit{96.89}                               & \textit{96.43}                            & \textit{96.21}                              \\ \cline{3-7} 
\multicolumn{1}{|l|}{}                                    & \multicolumn{1}{l|}{}                                    & \textit{10\%}                                                                         & \textit{94.23}                              & \textit{94.98}                               & \textit{94.71}                            & \textit{94.89}                              \\ \hline
\multicolumn{1}{|l|}{\multirow{3}{*}{\textit{SIRI-WHU}}}  & \multicolumn{1}{l|}{\multirow{3}{*}{\textit{MLRSNet}}}   & \textit{100\%}                                                                        & \textit{97.87}                              & \textit{97.43}                               & \textit{97.54}                            & \textit{97.32}                              \\ \cline{3-7} 
\multicolumn{1}{|l|}{}                                    & \multicolumn{1}{l|}{}                                    & \textit{50\%}                                                                         & \textit{94.40}                              & \textit{94.87}                               & \textit{94.91}                            & \textit{94.51}                              \\ \cline{3-7} 
\multicolumn{1}{|l|}{}                                    & \multicolumn{1}{l|}{}                                    & \textit{10\%}                                                                         & \textit{90.02}                              & \textit{90.83}                               & \textit{90.26}                            & \textit{90.73}                              \\ \hline
\end{tabular}%
}
\end{table}

\subsection{Multi-class Classification}
For the multi-class classification task, three different datasets have been used to evaluate contrastive learning for satellite image classification. The datasets are UC Merced, which has 21 classes on which our model achieved an accuracy of 99.35\%, precision of 99.91\%, recall value of 98.95\%, and F1 score of 98.89\% on the 100\% dataset in the downstream task. Another dataset used is the SIRI-WHU with 12 classes, and our model scored an accuracy of 99.68\%, precision of 95.36\%, recall value of 96.56\%, and F1 score of 97.92\% on the 100\% dataset in the downstream. MLRSNET is another dataset that was used, a 46 class dataset and our model achieved an accuracy of 96.59\%, precision of 96.79\%, recall value of 96.545, and F1 score of 96.65\% with 100\% data in the downstream. The results of further experiments are shown in Table \ref{tab4}, with fewer datasets downstream.

% \usepackage{multirow}
% \usepackage{graphicx}
\begin{table}[htbp]
\caption{Results for Multiclassification using contrastive learning}
\label{tab4}
\resizebox{\columnwidth}{!}{%
\begin{tabular}{|lll|c|c|c|c|}
\hline
\multicolumn{3}{|c|}{\textit{\textbf{Dataset used}}}                                                                                                                                                        & \multirow{2}{*}{\textit{\textbf{Accuracy}}} & \multirow{2}{*}{\textit{\textbf{Precision}}} & \multirow{2}{*}{\textit{\textbf{Recall}}} & \multirow{2}{*}{\textit{\textbf{F1-Score}}} \\ \cline{1-3}
\multicolumn{1}{|l|}{\textit{\textbf{Pretext}}}           & \multicolumn{1}{l|}{\textit{\textbf{Downstream}}}        & \textit{\textbf{\begin{tabular}[c]{@{}l@{}}\% of data \\ in downstream\end{tabular}}} &                                             &                                              &                                           &                                             \\ \hline
\multicolumn{1}{|l|}{\multirow{3}{*}{\textit{UC Merced}}} & \multicolumn{1}{l|}{\multirow{3}{*}{\textit{UC Merced}}} & \textit{100\%}                                                                        & \textit{99.35}                              & \textit{99.91}                               & \textit{98.95}                            & \textit{98.89}                              \\ \cline{3-7} 
\multicolumn{1}{|l|}{}                                    & \multicolumn{1}{l|}{}                                    & \textit{50\%}                                                                         & \textit{95.98}                              & \textit{91.78}                               & \textit{93.32}                            & \textit{93.82}                              \\ \cline{3-7} 
\multicolumn{1}{|l|}{}                                    & \multicolumn{1}{l|}{}                                    & \textit{10\%}                                                                         & \textit{89.67}                              & \textit{87.63}                               & \textit{85.86}                            & \textit{85.85}                              \\ \hline
\multicolumn{1}{|l|}{\multirow{3}{*}{\textit{SIRI-WHU}}}  & \multicolumn{1}{l|}{\multirow{3}{*}{\textit{SIRI-WHU}}}  & \textit{100\%}                                                                        & \textit{99.68}                              & \textit{95.36}                               & \textit{96.56}                            & \textit{97.92}                              \\ \cline{3-7} 
\multicolumn{1}{|l|}{}                                    & \multicolumn{1}{l|}{}                                    & \textit{50\%}                                                                         & \textit{95.54}                              & \textit{94.47}                               & \textit{95.29}                            & \textit{94.78}                              \\ \cline{3-7} 
\multicolumn{1}{|l|}{}                                    & \multicolumn{1}{l|}{}                                    & \textit{10\%}                                                                         & \textit{88.43}                              & \textit{83.76}                               & \textit{79.77}                            & \textit{81.43}                              \\ \hline
\multicolumn{1}{|l|}{\multirow{3}{*}{\textit{MLRSNet}}}   & \multicolumn{1}{l|}{\multirow{3}{*}{\textit{MLRSNet}}}   & \textit{100\%}                                                                        & \textit{96.59}                              & \textit{96.79}                               & \textit{96.54}                            & \textit{96.65}                              \\ \cline{3-7} 
\multicolumn{1}{|l|}{}                                    & \multicolumn{1}{l|}{}                                    & \textit{50\%}                                                                         & \textit{93.87}                              & \textit{91.37}                               & \textit{91.88}                            & \textit{92.68}                              \\ \cline{3-7} 
\multicolumn{1}{|l|}{}                                    & \multicolumn{1}{l|}{}                                    & \textit{10\%}                                                                         & \textit{88.66}                              & \textit{88.54}                               & \textit{88.91}                            & \textit{88.44}                              \\ \hline
\end{tabular}%
}
\end{table}

\subsection{Comparison with existing results}
The concept of self-supervised learning and domain adaptation-based self-supervised learning applied to satellite imagery has yet to be explored. This work considered various methods applied to these datasets, including supervised learning-based methods. Table 6 compares the results obtained from the proposed work with the existing binary and multi-class classification methods on satellite imagery.

% \usepackage{multirow}
% \usepackage{graphicx}
\begin{table}[htbp]
\caption{Comparative results of multi-class classification 21 class UCMD. (Top 2  results are shown)}
\label{tabc1}
\resizebox{\columnwidth}{!}{%
\begin{tabular}{|c|c|c|}
\hline
\textit{\textbf{Author}}                       & \textit{\textbf{Method}}                              & \textit{\textbf{Accuracy}}                                     \\ \hline
\textit{\cite{b23}}                           & \textit{SVM}                                          & \textit{\textbf{98.8}}                                                  \\ \hline
\textit{\cite{b24}}                       & \textit{GIST}                                         & \textit{46.9}                                                  \\ \hline
\textit{\cite{b25}}                           & \textit{ResNet 50}                                    & \textit{98}                                                    \\ \hline
\textit{\cite{b26}}                    & \textit{DCNN}                                         & \textit{93.48}                                                 \\ \hline
\textit{\cite{b16}}                  & \textit{GoogleNet}                                    & \textit{\begin{tabular}[c]{@{}c@{}} 97.10\end{tabular}} \\ \hline
\textit{\cite{b14}}                          & \textit{Semisupervised ensemble projection}           & \textit{66.49}                                                 \\ \hline
\multirow{2}{*}{\textit{\textbf{Our Results}}} & \textit{\textbf{Self-supervised Domain Adaptation}}   & \textit{98.75}                                        \\ \cline{2-3} 
                                               & \textit{\textbf{Self-supervised Same Dataset}} & \textit{\textbf{99.35}}                                        \\ \hline
\end{tabular}%
}
\end{table}

\begin{table}[htbp]
\caption{Comparative results for multi-class classification 12 class SIRI-WHU. (Top 2  results are shown)}
\label{tabc2}
\resizebox{\columnwidth}{!}{%
\begin{tabular}{|c|c|c|}
\hline
\textit{\textbf{Author}}                       & \textit{\textbf{Method}}                              & \textit{\textbf{Accuracy}} \\ \hline
\textit{\cite{b27}}                            & \textit{AlexNet SPP SS}                               & \textit{95.07}             \\ \hline
\textit{\cite{b28}}                            & \textit{MCNN}                                         & \textit{93.75}             \\ \hline
\textit{\cite{b29}}                            & \textit{Inception-LSTM}                               & \textit{\textbf{99.73}}             \\ \hline
\multirow{2}{*}{\textit{\textbf{Our Results}}} & \textit{\textbf{Self-Supervised Domain Adaptation}}   & \textit{96.87}    \\ \cline{2-3} 
                                               & \textit{\textbf{Self-supervised Same Dataset}} & \textit{\textbf{99.68}}    \\ \hline
\end{tabular}%
}
\end{table}

\begin{table}[H]
\caption{Comparative results for multi-class classification 46 class MLRSNet. (Top 2  results are shown)}
\label{tabc3}
\resizebox{\columnwidth}{!}{%
\begin{tabular}{|c|c|c|}
\hline
\textit{\textbf{Author}}                       & \textit{\textbf{Method}}                              & \textit{\textbf{Accuracy}} \\ \hline
\textit{\cite{b30}}                             & \textit{DenseNet201-SR-Net}                           & \textit{87.87}             \\ \hline
\textit{\cite{b30}}                             & \textit{ResNet101-SR-Net}                             & \textit{87.48}             \\ \hline
\textit{\cite{b17}}                             & \textit{Self-Supervised Learning}                     & \textit{96}                \\ \hline
\multirow{2}{*}{\textit{\textbf{Our Results}}} & \textit{\textbf{Self-Supervised Domain Adaptation}}   & \textit{\textbf{97.87}}    \\ \cline{2-3} 
                                               & \textit{\textbf{Self-supervised Same Dataset}} & \textit{\textbf{96.59}}    \\ \hline
\end{tabular}%
}
\end{table}

Based on the comparisons in Tables \ref{tabc1}, \ref{tabc2}, \ref{tabc3} the proposed work performs better than the existing works. 
According to the above comparative analysis, the proposed work outperforms all previous works and achieves state-of-the-art results for multi-class classification of satellite imagery.

\section{Discussions}
This section discusses the key outcomes of the proposed work and provides analysis based on the achieved results on three datasets for different scenarios.
\subsection{Self-supervised learnt representations are domain adaptable}
\begin{figure}[htbp]
\centerline{
\includegraphics[width=8cm, height=7cm]{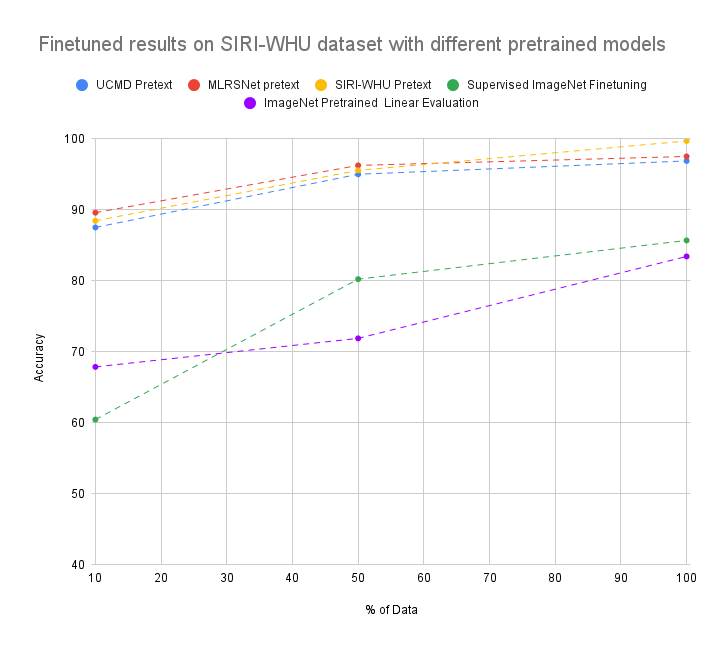} 
}
\caption{Comparison of accuracy of the proposed method for SIRI-WHU with supervised learning.}
\label{g1}
\end{figure}

\begin{figure}[htbp]
\centerline{
\includegraphics[width=8cm, height=7cm]{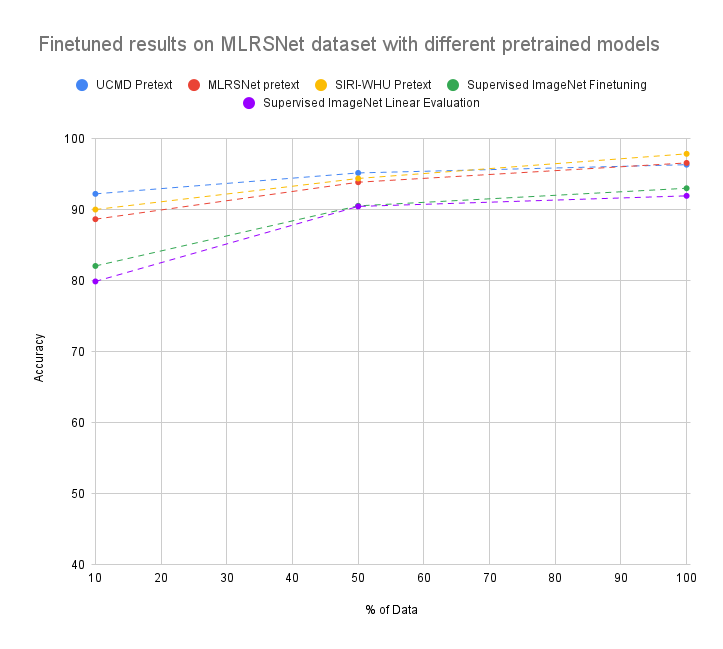} 
}
\caption{Comparison of accuracy of the proposed method for MLRSNet with supervised learning.}
\label{g2}
\end{figure}

\begin{figure}[htbp]
\centerline{
\includegraphics[width=8cm, height=7cm]{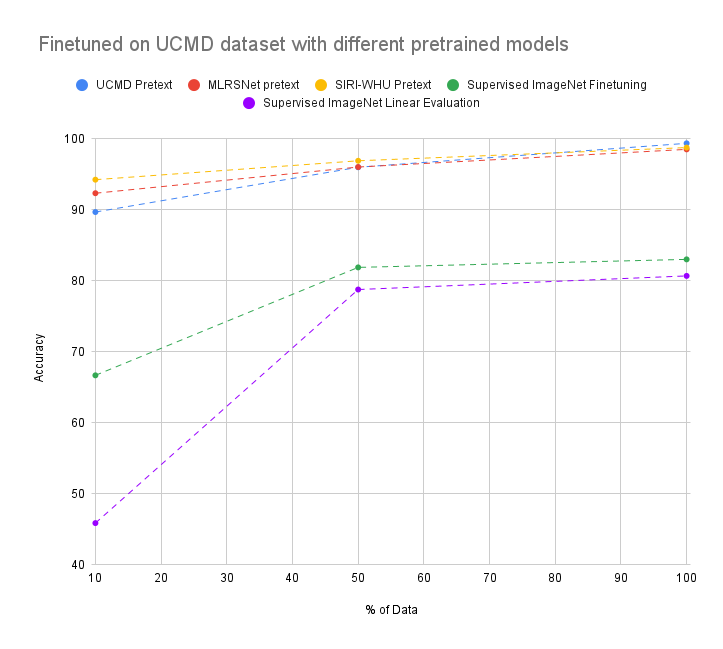} 
}
\caption{Comparison of accuracy of the proposed method for UCMD with supervised learning.}
\label{g3}
\end{figure}

Results in Table~\ref{tab3} clearly indicate that the performance of domain adaptation with different sources and targets achieves comparable results with in-domain knowledge transfer presented in Table~\ref{tab4}. While investigating and comparing the domain adaptation results with ImageNet supervised knowledge transfer (refer Table~\ref{ft} \& \ref{le}), all the models outperform which indicates the successful domain adaptations across the datasets.
Following the trend, the proposed framework consistently outperformed on the given datasets compared with the ImageNet pretrained ResNet50 in a complete range of labels from 10\% to 100\%, shown in Figure \ref{g1}, \ref{g2}, \& \ref{g3}.

\begin{table}[htbp]
\caption{Results for ResNet50 finetuning with Imagenet weights}
\label{ft}
\resizebox{\columnwidth}{!}{%
\begin{tabular}{|l|l|c|c|c|c|c|}
\hline
\textit{\textbf{Dataset}}          & \textit{\textbf{\% of data}} & \textit{\textbf{Accuracy}} & \textit{\textbf{Precision}} & \textit{\textbf{Recall}} & \textit{\textbf{F1-Score}} & \textit{\textbf{AUC}} \\ \hline
\multirow{3}{*}{\textit{UCMD}}     & \textit{100\%}               & \textit{83.00}             & \textit{82.89}              & \textit{82.34}           & \textit{82.47}             & \textit{93.83}        \\ \cline{2-7} 
                                   & \textit{50\%}                & \textit{81.87}             & \textit{81.77}              & \textit{81.32}           & \textit{81.53}             & \textit{92.93}        \\ \cline{2-7} 
                                   & \textit{10\%}                & \textit{66.66}             & \textit{65.78}              & \textit{66.45}           & \textit{66.21}             & \textit{88.87}        \\ \hline
\multirow{3}{*}{\textit{SIRI-WHU}} & \textit{100\%}               & \textit{85.67}             & \textit{85.54}              & \textit{85.33}           & \textit{85.44}             & \textit{95.99}        \\ \cline{2-7} 
                                   & \textit{50\%}                & \textit{80.21}             & \textit{80.02}              & \textit{79.99}           & \textit{79.34}             & \textit{93.95}        \\ \cline{2-7} 
                                   & \textit{10\%}                & \textit{60.71}             & \textit{60.43}              & \textit{60.33}           & \textit{60.21}             & \textit{90.99}        \\ \hline
\multirow{3}{*}{\textit{MLRSNet}}  & \textit{100\%}               & \textit{93.02}             & \textit{92.98}              & \textit{92.74}           & \textit{92.34}             & \textit{97.99}        \\ \cline{2-7} 
                                   & \textit{50\%}                & \textit{90.54}             & \textit{90.32}              & \textit{90.55}           & \textit{90.43}             & \textit{96.99}        \\ \cline{2-7} 
                                   & \textit{10\%}                & \textit{82.07}             & \textit{81.76}              & \textit{81.34}           & \textit{81.33}             & \textit{93.69}        \\ \hline
\end{tabular}%
}
\end{table}

\begin{table}[htbp]
\caption{Results for ResNet50 linear evaluation with Imagenet weights}
\label{le}
\resizebox{\columnwidth}{!}{%
\begin{tabular}{|l|l|c|c|c|c|c|}
\hline
\textit{\textbf{Dataset}}          & \textit{\textbf{\% of data}} & \textit{\textbf{Accuracy}} & \textit{\textbf{Precision}} & \textit{\textbf{Recall}} & \textit{\textbf{F1-Score}} & \textit{\textbf{AUC}} \\ \hline
\multirow{3}{*}{\textit{UCMD}}     & \textit{100\%}               & \textit{80.67}             & \textit{80.32}              & \textit{78.98}           & \textit{80.16}             & \textit{93.23}        \\ \cline{2-7} 
                                   & \textit{50\%}                & \textit{78.75}             & \textit{77.97}              & \textit{78.78}           & \textit{78.84}             & \textit{93.12}        \\ \cline{2-7} 
                                   & \textit{10\%}                & \textit{45.84}             & \textit{45.43}              & \textit{45.12}           & \textit{45.59}             & \textit{88.98}        \\ \hline
\multirow{3}{*}{\textit{SIRI-WHU}} & \textit{100\%}               & \textit{83.41}             & \textit{83.78}              & \textit{83.43}           & \textit{83.61}             & \textit{93.99}        \\ \cline{2-7} 
                                   & \textit{50\%}                & \textit{71.87}             & \textit{71.43}              & \textit{71.91}           & \textit{71.79}             & \textit{92.87}        \\ \cline{2-7} 
                                   & \textit{10\%}                & \textit{67.85}             & \textit{67.99}              & \textit{67.65}           & \textit{67.31}             & \textit{90.54}        \\ \hline
\multirow{3}{*}{\textit{MLRSNet}}  & \textit{100\%}               & \textit{91.95}             & \textit{90.93}              & \textit{91.99}           & \textit{91.63}             & \textit{95.67}        \\ \cline{2-7} 
                                   & \textit{50\%}                & \textit{90.46}             & \textit{88.65}              & \textit{90.32}           & \textit{90.91}             & \textit{94.09}        \\ \cline{2-7} 
                                   & \textit{10\%}                & \textit{79.89}             & \textit{80.00}              & \textit{79.02}           & \textit{79.45}             & \textit{85.34}        \\ \hline
\end{tabular}%
}
\end{table}

\subsection{Self-supervised representation based knowledge transfer demonstrate label efficiency}
Results on all three datasets indicate that the proposed framework obtains state-of-the-art results with only $10\%$ and $50\%$ of labels comparing previous work, which indicates that self-supervised learnt representations capture the important features of visual concepts of interest and adapt it for downstream tasks without additional efforts and parameter adjustments. 
\subsection{Self-supervised pretrianed models outperforms in fully supervised setting}
Besides label efficiency, knowledge transfer in self-supervised pretrained models outperforms previous works for all three datasets. This trend indicates that self-supervised representations are efficient end-to-end learning.
\subsection{Robust and explainable representations}
\begin{figure}[htbp]
\centerline{
\includegraphics[width=9cm, height=9cm]{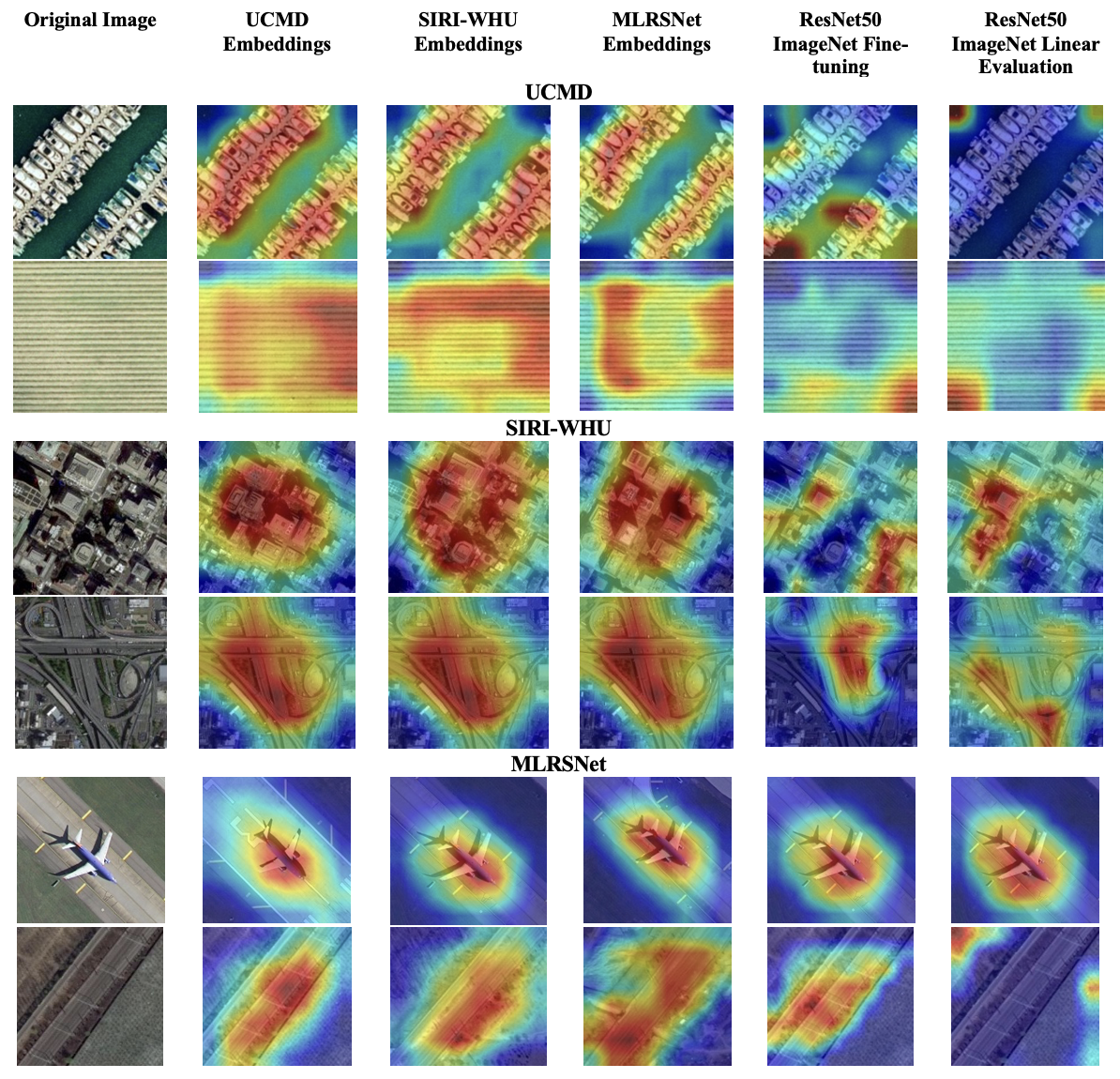} 
}
\caption{Learning representation of the architecture using Class Activation Maps }
\label{fig7}
\end{figure}

All the quantitative results are well described, with qualitative analysis performed on all three datasets with activation maps. Figure~\ref{fig7} demonstrates the explainability and attention for self-supervised pretrained models against ImageNet supervised models. This indicates that learnt representations in self-supervised manner are more efficient and thus achieve higher performance in downstream tasks. 

\section{Related Work}
During the past few years, self-supervised contrastive learning has emerged as a new training paradigm. Using this training paradigm, comprehensive representations can be learned without human annotation, which could solve the lack of annotated data problem. Much research has yet to be done on self-supervised learning in satellite imagery. Here, the main focus is on discussing deep learning methods applied to satellite imagery classification. 
\subsection{Supervised learning on satellite images}
R. Naushad et al. \cite{b13} proposed a transfer learning approach to classify land use and land cover on the Eurosat dataset. For this, four CNN models were pre-trained: VGG-16 (without data augmentation), VGG-16 (with data augmentation), wide ResNet-50 (Without Data Augmentation), and wide ResNet-50 (With Data Augmentation). They achieved an accuracy of 99.17\%. However, the proposed approach has been validated only on one dataset (the Eurosat dataset ) while using 100\% of the data. At a large scale, CNN-based models were used by A. Albert et al. \cite{b14} to identify patterns in urban environments using satellite imagery. They used pre-trained models: VGG 16 and ResNet, for the classification task and achieved different accuracies for different countries, with which authors showed the highest accuracy of 83\%. However, there remains scope for improvement in the results. Another method was applied by M. Castelluccio et al.  \cite{b16} for land use classification in remote sensing images. They used pre-trained models: CaffeNet and GoogLeNet. These models provided an accuracy of 90.17\% and 91.83\%, respectively. Although authors achieve competitive results, the possibility of improvement of results and the use of better models remains.\\
X. Tan et al. proposed a multilabel classification to classify the MLRSNet, a benchmark dataset of 46 classes.  They achieved an accuracy of 87.87\% with DenseNet201-SR-Net\cite{b30}. However, there is still scope for improvement in accuracy. Furthermore, DenseNet201 is a very heavy computational method that needs more resources. S.Jog et al. \cite{b19} performed a supervised classification of satellite images using the Landsat dataset and support vector machine as the classifier, which achieved an accuracy of 92.84\%. However, this approach was tested on a single dataset. Thus,  the proposed method needs to be validated on other datasets as well to check the robustness of the model. M. Pritt et al. \cite{b22} used convolutional neural networks on the FMoV dataset for satellite image classification and achieved an accuracy of 83\%. However, there is much scope for improvement in accuracy. \\
Özyurt, F. et al. \cite{b23} attempted to classify satellite images using the UC Merced dataset with a unique approach to feature extraction. For the classification, they used an SVM-based machine learning model and achieved an accuracy of 98.8\%. However, their work focussed on a single dataset only. Kadhim et al. \cite{b25} used a pre-trained ResNet50 model on the UC Merced dataset for the satellite image classification task and achieved  an accuracy of 98\%. However, the model has been evaluated on a single dataset only. F. P. S. Luus et. al. \cite{b26} used deep convolutional neural networks for land use classification on the UC Merced dataset and achieved an accuracy of 93.48\%. Though the methodology used differs from existing methods, there is a scope for improvement in accuracy.\\
Han. Xiaobing et.al. \cite{b27}  used a pre-trained AlexNet model on the SIRI-WHU dataset for satellite image classification and obtained an accuracy of 95.07\%. However, the accuracy can be improved further using better network architectures. Y. Liu et al. \cite{b28} worked on the SIRI-WHU dataset and used multiscale convolutional neural networks for scene classification using satellite images. They achieved an accuracy of 93.75\%. Although the method differs from existing approaches, this work is based on a single dataset only and needs to be tested on other datasets also. Y. Dong et al. \cite{b29} classified satellite images from the SIRI-WHU dataset using inception-based LSTM approach with an accuracy of 99.73\%. However, the approach is tested on a single dataset only. 
\subsection{Self-supervision and domain adaptation on specialized visual domain}
Self-supervised methods on ImageNet and natural scenes have advanced in recent times. It has also been considerable advancements in other specialized visual domains to adapt self-supervised representation learning. Self-supervised methods in medical images showed progress where data and human labels are limited, Chhipa et al.~\cite{b33} demonstrated self-supervised domain adaptation on histopathology images, and Azizi et al.~\cite{b34} showed knowledge transfer on X-ray. Other interesting applications for self-supervised methods exploring on underwater images Tarling et al.~\cite{b37} for the fish count and identifying mining materials from three-dimensional particle management sensors in Chhipa et al.~\cite{b36} shown progress in a specialized domain.   
\subsection{Domain adaptation and self-supervised learning on satellite images}
Few non-supervised learning based methods for satellite image classification have been proposed in the literature.  A semi-supervised learning based approach for satellite image classification was proposed by W. Yang et al. \cite{b15} to solve the problem of fewer images. They achieved an accuracy of 73.82\% on 19 class data and 65.34\% on UCMD dataset. However, there is scope for further improvements in the results obtained by the authors  for  satellite image classification.\\
A few self-supervised methods have also been used to classify remotely sensed satellite images in the literature, as proposed by V. Stojnic et al. \cite{b17}. They used self-supervised methods with a pre-trained Imagenet model on MLRSNet and achieved an accuracy of 96\%. Manas et al.~\cite{b35} have shown self-supervised pretraining on remote sensing data using weather information.
%However, their approach must be validated on different datasets and cross-domain settings.
Yi Wang et al. \cite{b18} proposed contrastive multiview coding (CMC) based approach for satellite image classification, where one image is an anchor, and other images are neighboured around that image. They used pre-trained models for feature extraction, and the number of training samples was large. However, they did not validate the proposed approach in cross-domain settings wherein learning the representations from one dataset of satellite images and performing downstream tasks on another dataset.\\
From the above analysis of the existing work in the literature, it can easily be observed that most of the existing supervised learning-based satellite image classification methods require a lot of labeled data to perform satisfactorily. Only a few semi-supervised or self-supervised satellite image classification methods exist in the literature. However, these methods use the same dataset for pretext tasks, and downstream tasks, and these methods have not been evaluated in cross-domain settings. To mitigate these research gaps in the literature, this work proposes a domain adaptation-based self-supervised representation learning approach for classifying satellite images. This work proposes a domain adaptable self-supervised learning approach to reuse the representations learned on one unlabelled dataset from the source domain for classifying satellite images taken from a different target domain dataset.

\section{Conclusion}
%Classifying remotely sensed satellite images is a task that involves interpreting and inferring results to gain a better understanding of the developments that have been made or will be made in the future. 
This work proposed a domain-adaptable self-supervised representation learning based framework focusing on the robust evaluation of learnt representations rather than one-directional knowledge transfer, which ultimately reviews the effectiveness and applicability of such methods in the satellite imagery visual domain.
One significant outcome is achieving improved performance by applying domain-adapted knowledge transfer across the datasets, outperforming the existing methods of satellite image classification, even in cross-domain settings. By applying the self-supervised representation learning, the proposed work has surpassed the existing results by ~1\%, with fewer training data.
The proposed evaluation framework is conveniently applicable to other visual domains which are not thoroughly explored yet for the usability of self-supervised representation learning to reduce human annotation needs.
In future work, i) We aim to investigate domain adaptation for other computer vision downstream tasks, i.e., segmentation and localization, ii) Investigate non-contrastive representation learning methods, and iii) Candidates for standard augmentation methods in self-supervised learning to adapt remote sensing visual domain.

\vspace{12pt}

\end{document}